\documentclass{article}

    \usepackage[preprint]{paper_style}

\usepackage[utf8]{inputenc} 
\usepackage[T1]{fontenc}    
\usepackage{hyperref}       
\usepackage{url}            
\usepackage{booktabs}       
\usepackage{amsfonts}       
\usepackage{nicefrac}       
\usepackage{microtype}      
\usepackage{xcolor}         

\usepackage{graphicx} 
\usepackage{adjustbox}
\usepackage{comment}
\usepackage{enumitem}
\usepackage{amsmath} 

\title{RLBenchNet: The Right Network for the Right Reinforcement Learning Task}

\author{%
  Ivan Smirnov \\
  Deggendorf Institute of Technology\\  
  \And
  Shangding Gu \thanks{corresponding to \textit{shangding.gu@cs.tum.edu}} \\
  Technical University of Munich
}

\begin{document}

\maketitle

\begin{abstract}
Reinforcement learning (RL) has seen significant advancements through the application of various neural network architectures. In this study, we systematically investigate the performance of several neural networks in RL tasks, including Long Short-Term Memory (LSTM), Multi-Layer Perceptron (MLP), Mamba/Mamba-2, Transformer-XL, Gated Transformer-XL, and Gated Recurrent Unit (GRU). Through comprehensive evaluation across continuous control, discrete decision-making, and memory-based environments, we identify architecture-specific strengths and limitations. Our results reveal that: (1) MLPs excel in fully observable continuous control tasks, providing an optimal balance of performance and efficiency; (2) recurrent architectures like LSTM and GRU offer robust performance in partially observable environments with moderate memory requirements; (3) Mamba models achieve a 4.5× higher throughput compared to LSTM and a 3.9× increase over GRU, all while maintaining comparable performance; and (4) only Transformer-XL, Gated Transformer-XL, and Mamba-2 successfully solve the most challenging memory-intensive tasks, with Mamba-2 requiring 8× less memory than Transformer-XL. These findings provide insights for researchers and practitioners, enabling more informed architecture selection based on specific task characteristics and computational constraints. Code is available at: \url{https://github.com/SafeRL-Lab/RLBenchNet}
\end{abstract}

\section{Introduction}

Reinforcement learning (RL) has emerged as a powerful framework for sequential decision-making, with neural networks playing a key role in enabling agents to learn complex policies \cite{gu2023safe, gu2025safe, silver2016mastering}. Despite their importance, the influence of neural network architecture on RL performance across diverse environments remains relatively underexplored in the literature.
In this work, we aim to bridge this gap by systematically evaluating the impact of various neural network architectures on the performance of RL agents. 

To ground our investigation, we choose Proximal Policy Optimization (PPO) \cite{schulman2017proximalpolicyoptimizationalgorithms} as the primary algorithm for our study, as it is one of the most widely adopted reinforcement learning methods, known for its simplicity and strong empirical performance. Our experiments span a diverse set of benchmark tasks, offering comprehensive insights into how architectural choices affect learning efficiency and policy quality in RL. These tasks include environments requiring memory, such as Partially Observable Markov Decision Processes (POMDPs), and environments focused on continuous control and discrete decision-making. By analyzing the strengths and weaknesses of architectures such as LSTM \cite{hochreiter1997long}, GRU \cite{chung2014empiricalevaluationgatedrecurrent}, Transformer-XL \cite{dai2019transformerxlattentivelanguagemodels},  Gated Transformer-XL (GTrXL) \cite{parisotto2019stabilizingtransformersreinforcementlearning}, Mamba \cite{gu2024mambalineartimesequencemodeling}, Mamba-2 \cite{dao2024transformersssmsgeneralizedmodels} and MLP, we aim to provide actionable insights into the design of RL systems.

Previous works have explored implementation details of PPO \cite{shengyi2022the37implementation}, that methods could be used to improve the agent performance in various environments \cite{andrychowicz2020mattersonpolicyreinforcementlearning}, and studies like \cite{pleines2024memorygymendlesstasks} have demonstrated the efficacy of Transformer-XL in episodic memory tasks. However, these studies often overlook comparisons with simpler architectures, such as PPO-LSTM, and emerging architectures, like PPO-Mamba. Furthermore, existing benchmarks, such as those conducted in Memory Gym \cite{pleines2024memorygymendlesstasks} and MiniGrid \cite{chevalierboisvert2023minigridminiworldmodular}, highlight the need for memory in certain environments but lack comprehensive comparisons across a broader range of architectures and tasks.

Our contributions are threefold:
\begin{itemize}[leftmargin=*]
    \item \textbf{Architecture Benchmarking in RL:} We benchmark RL (PPO) implementations using a variety of neural network architectures, including traditional models (MLP, LSTM, GRU), advanced architectures (Transformer-XL, GTrXL), and novel designs (Mamba, Mamba-2). We also provide insights into selecting appropriate neural networks for different settings.
    \item \textbf{Evaluation Across Diverse Tasks:} We evaluate these neural network architectures across memory-intensive environments such as MiniGrid; partially observable classical control tasks, including LunarLander and CartPole; and both continuous and discrete control domains, such as MuJoCo \cite{todorov2012mujoco} and Atari \cite{Bellemare_2013}. We further provide a comprehensive analysis of the experimental results.
    \item \textbf{Trade-off and Guideline Analysis:} We analyze the trade-offs between memory requirements, computational efficiency, and task performance, offering practical guidelines for selecting architectures based on task characteristics.
\end{itemize}

\section{Related Work}
\subsection{Memory Models for Reinforcement Learning}
Memory modeling in RL has evolved through three main architectural paradigms: \textit{recurrent networks}, \textit{transformer-based models}, and \textit{state-space models}. Each of these architectures offers distinct advantages depending on the task characteristics, particularly in environments with varying levels of partial observability and memory requirements.

\textit{Recurrent networks}, including LSTM and GRU \cite{hochreiter1997long, chung2014empiricalevaluationgatedrecurrent}, have been the traditional choice for handling partial observability in RL. These architectures maintain an internal memory state that is updated at each timestep, enabling them to integrate information over time. Studies like MemoryGym \cite{pleines2024memorygymendlesstasks} have shown that well-tuned GRU models can outperform even advanced architectures like Transformer-XL in indefinite-horizon tasks \cite{pleines2023trxlppo}. However, their effectiveness is limited in long-horizon scenarios due to vanishing gradient issues and fixed memory capacity \cite{lu2024rethinkingtransformerssolvingpomdps}.

\textit{Transformer-based architectures}, such as Transformer-XL and GTrXL \cite{dai2019transformerxlattentivelanguagemodels, parisotto2019stabilizingtransformersreinforcementlearning}, introduce self-attention mechanisms that excel at modeling long-term dependencies. These models have demonstrated state-of-the-art performance in partially observable environments \cite{ni2023transformersshinerldecoupling}, but their quadratic complexity with sequence length can become a computational bottleneck. Hybrid models, like the Recurrent Memory Transformer (RMT) \cite{cherepanov2024recurrentactiontransformermemory} and Recurrent Action Transformer with Memory (RATE) \cite{bulatov2024scalingtransformer1mtokens}, extend transformer capabilities by integrating external memory mechanisms.

\textit{State-space models}, particularly Mamba and Mamba-2 \cite{gu2024mambalineartimesequencemodeling, dao2024transformersssmsgeneralizedmodels}, represent a recent innovation, balancing the computational efficiency of recurrent networks with the expressive capacity of transformers. Mamba employs a selective state-space mechanism that efficiently captures temporal dependencies without the overhead of self-attention. Mamba-2 further improves long-horizon memory retention, making it competitive even in highly memory-dependent tasks. Recent work has integrated Mamba into decision-making frameworks, such as Decision Mamba \cite{lv2025decisionmambamultigrainedstate, ota2024decisionmambareinforcementlearning}, highlighting its adaptability across tasks.

These three architectural paradigms offer distinct strengths: recurrent networks provide robust solutions for short-term memory tasks, transformers excel in complex, long-horizon scenarios, and state-space models like Mamba achieve an optimal balance of computational efficiency and performance. Our work systematically benchmarks these architectures in RL, revealing their strengths, limitations, and suitable application domains.

\subsection{Benchmarking Neural Architectures in RL}

Recent years have seen increased interest in systematically evaluating different neural architectures across diverse RL tasks. Benchmark suites like POPGym \cite{morad2023popgymbenchmarkingpartiallyobservable} and Memory Gym \cite{pleines2024memorygymendlesstasks} have been developed specifically to assess how different memory models perform under varying degrees of partial observability. These benchmarks have revealed that no single architecture dominates across all tasks. While transformers excel when long but finite context is needed, recurrent networks often have an edge in continuously evolving tasks or when the agent must generalize from limited training data. Hybrid approaches that combine the strengths of different architectures (such as the Decision Mamba-Hybrid \cite{huang2024decisionmambareinforcementlearning} that uses Mamba for long-term memory and a transformer for short-term decision-making) have shown particular promise in complex environments.

Our work provides a systematic comparison of PPO implementations across diverse neural architectures (MLP, LSTM, GRU, Transformer-XL, GTrXL, Mamba, and Mamba-2) within a unified framework. By focusing solely on these base architectures without external memory augmentation, we offer a clear analysis of their intrinsic memory capabilities, enabling a direct evaluation of the trade-offs in performance and computational efficiency across various environments.

\section{Benchmarking Settings}

\subsection{Neural Network Architecture in RL}
To ensure consistency and reproducibility, we based our implementations on CleanRL \cite{huang2022cleanrl}, a popular open-source library for RL algorithms. Our approach maintains the core PPO algorithm while varying only the neural network architecture, allowing for fair comparisons focused on architectural differences rather than algorithmic variations. The following neural networks of RL that we benchmark in our study:

(1) \textbf{MLP:} Standard multi-layer perceptron with ReLU activations, serving as our baseline. In the baseline configuration, agents received only the current observation at each timestep, referred to as \textit{PPO-1} or \textit{MLP Obs. Stack 1}. To introduce a simple memory mechanism, we experimented with \textit{PPO-4}, also referred as \textit{MLP Obs. Stack 4}, where four consecutive observations were concatenated and fed into the network. For Atari, this follows the standard practice of frame stacking the last four frames. (2) \textbf{LSTM and GRU:} Recurrent networks implemented following CleanRL's PPO-LSTM structure, using a single recurrent layer. While this single-layer approach is consistent with typical implementations, it may have limitations in highly memory-dependent tasks. (3) \textbf{Transformer-XL, GTrXL:} Implemented using CleanRL's episodic memory structure for hidden states, which maintains information across episode boundaries through its segmented recurrent mechanism. This implementation also employs post-transformer MLP layers. The gated version introduces a learned gating mechanism in place of traditional skip connections. (4) 
\textbf{Mamba/Mamba-2:} Integrated using the official implementation from the \texttt{mamba-ssm} repository. For Mamba, we employed an optimized training approach utilizing the selective scan mechanism without resetting at episode boundaries, offering computational advantages but potentially introducing state leakage between episodes. We incorporated post-model MLP layers and layer normalization.

Particularly, for all architectures, we adjusted network sizes to maintain approximately equal parameter counts (see Table~\ref{tab:env_params_k} in Appendix), ensuring that performance differences reflect architectural capabilities rather than capacity disparities. Moreover, we provide single-file implementations for each architecture variant. This approach makes architectural differences explicit while maintaining consistent handling of environment interactions, data collection, and policy updates.

\subsection{Training Protocols}
We established consistent training protocols across environments while adjusting for their varying complexity, as detailed in Table ~\ref{tab:env_settings}.

For all environments, we used the default maximum episode length as specified in their standard implementations. In MiniGrid tasks, we further increased the challenge by reducing the agent's observation window from the default 7×7 to 3×3, making the environments more partially observable and thus more reliant on memory-based architectures.

\subsection{Hyperparameters and Hardware}
We primarily adhered to CleanRL's default PPO hyperparameters, with key parameters shown in Table~\ref {tab:common_hyperparams} from the Appendix. The only notable exception was the learning rate for Mamba-based models, which was reduced following recommendations from recent literature \cite{luo2024efficient} to enhance stability.

For \textbf{Mamba}, we used the default hyperparameter settings provided in the official implementation, as the available configuration space is relatively narrow and recent work suggests strong performance without extensive tuning.

For \textbf{Transformer-based models}, we utilized recommended hyperparameters for environments where prior work was available (e.g., MiniGrid, MemoryGym). In environments lacking specific benchmarks, we adjusted settings based on the environment’s complexity, particularly whether it requires short-term or long-term memory, without engaging in extensive hyperparameter tuning.

All experiments were conducted on a single NVIDIA RTX A5000 GPU and Intel Xeon W-1390p to ensure consistent performance measurement. Training throughput, inference latency, and memory usage were measured using PyTorch's built-in profiling tools.

\section{Results and Analysis}

We evaluate the performance of each architecture across diverse environments with varying requirements for memory, continuous control, and discrete decision-making.

\subsection{Continuous Control Tasks}

To systematically evaluate the impact of neural architectures on continuous control tasks, we conducted benchmarks on three popular MuJoCo environments: Walker2d-v4, HalfCheetah-v4, and Hopper-v4. These tasks present distinct challenges, ranging from stability-focused control (Hopper) to speed and efficiency (HalfCheetah). Our results demonstrate that the optimal architecture is highly task-dependent, reflecting the varying dynamics and control complexities of each environment.

\begin{center}
\fcolorbox{black}{blue!10}{\parbox{.98\linewidth}{\textcolor{magenta}{Finding}: 
\begin{itemize}[leftmargin=*, itemsep=0.5em]
    \item Mamba has the worst performance across all environments, while Mamba-2 shows significant improvement, with performance comparable to LSTM and GRU while training 5× faster. \vspace{-8pt}
    \item MLP performs well in most tasks, except for Hopper, where short memory capabilities are critical for maintaining stability.
\end{itemize}
}}
\end{center}

Specifically, our analysis reveals that environments with simpler, smooth dynamics (like HalfCheetah) are effectively modeled by feed-forward architectures (MLP), which achieve high performance with strong sample efficiency. In contrast, environments with higher stability demands (like Hopper) benefit from recurrent models (GRU, LSTM), which effectively capture temporal dependencies critical for maintaining balance. Notably, in Walker2d, MLPs again perform competitively, highlighting the value of simplicity when the task dynamics are less chaotic.

\begin{figure}[t]
\centering
\includegraphics[width=0.32\columnwidth]{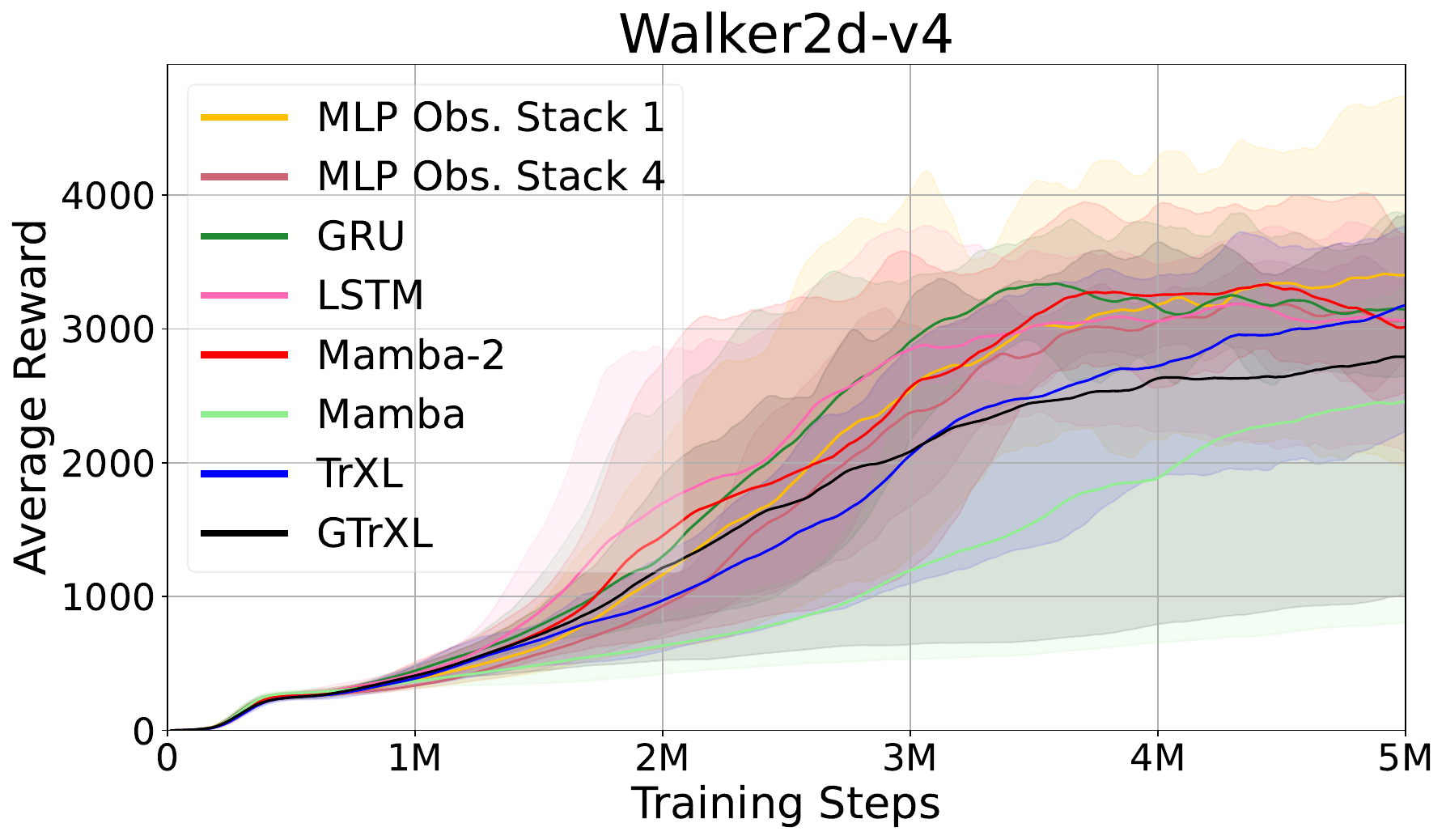}
\includegraphics[width=0.32\columnwidth]{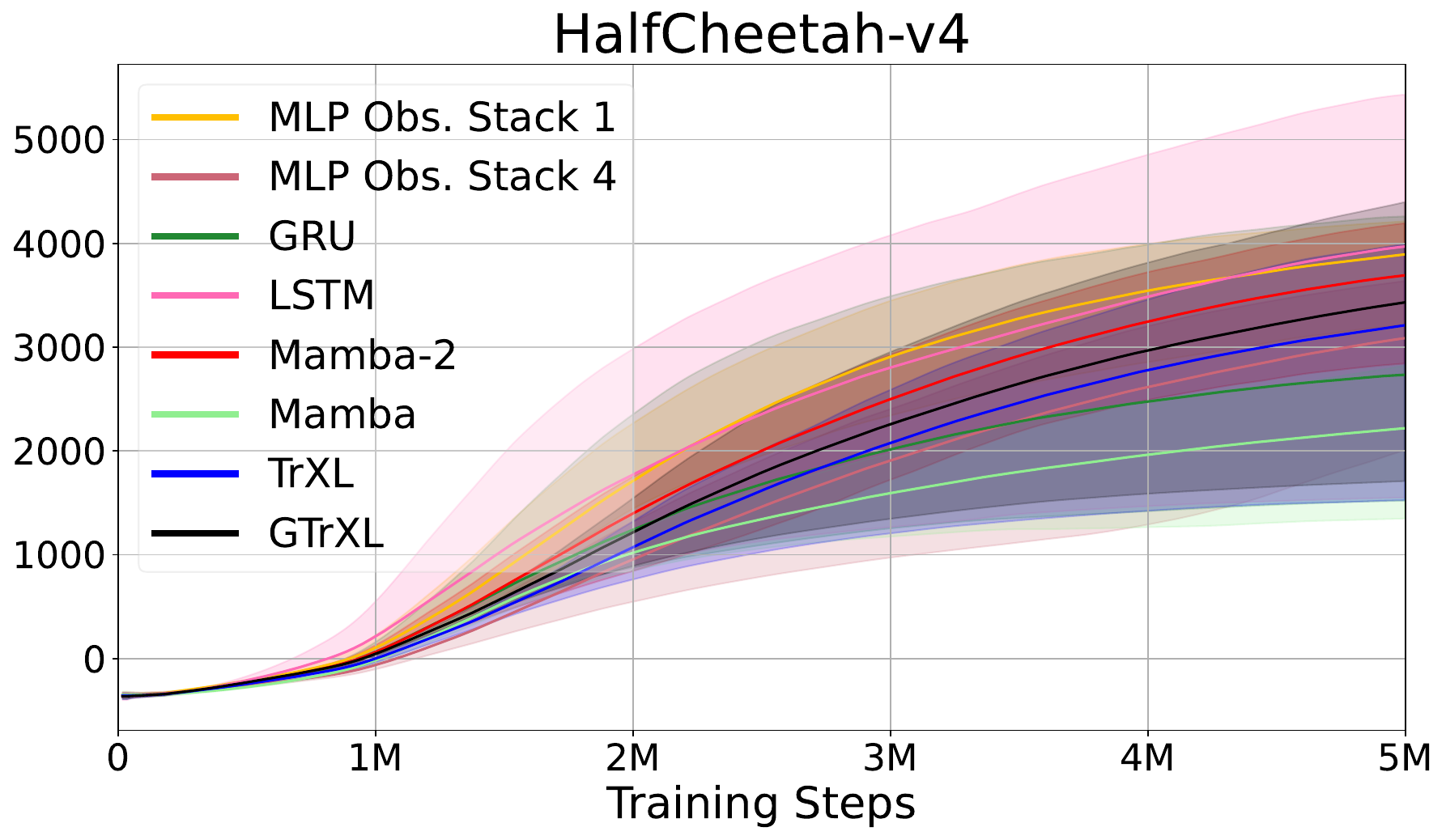}
\includegraphics[width=0.32\columnwidth]{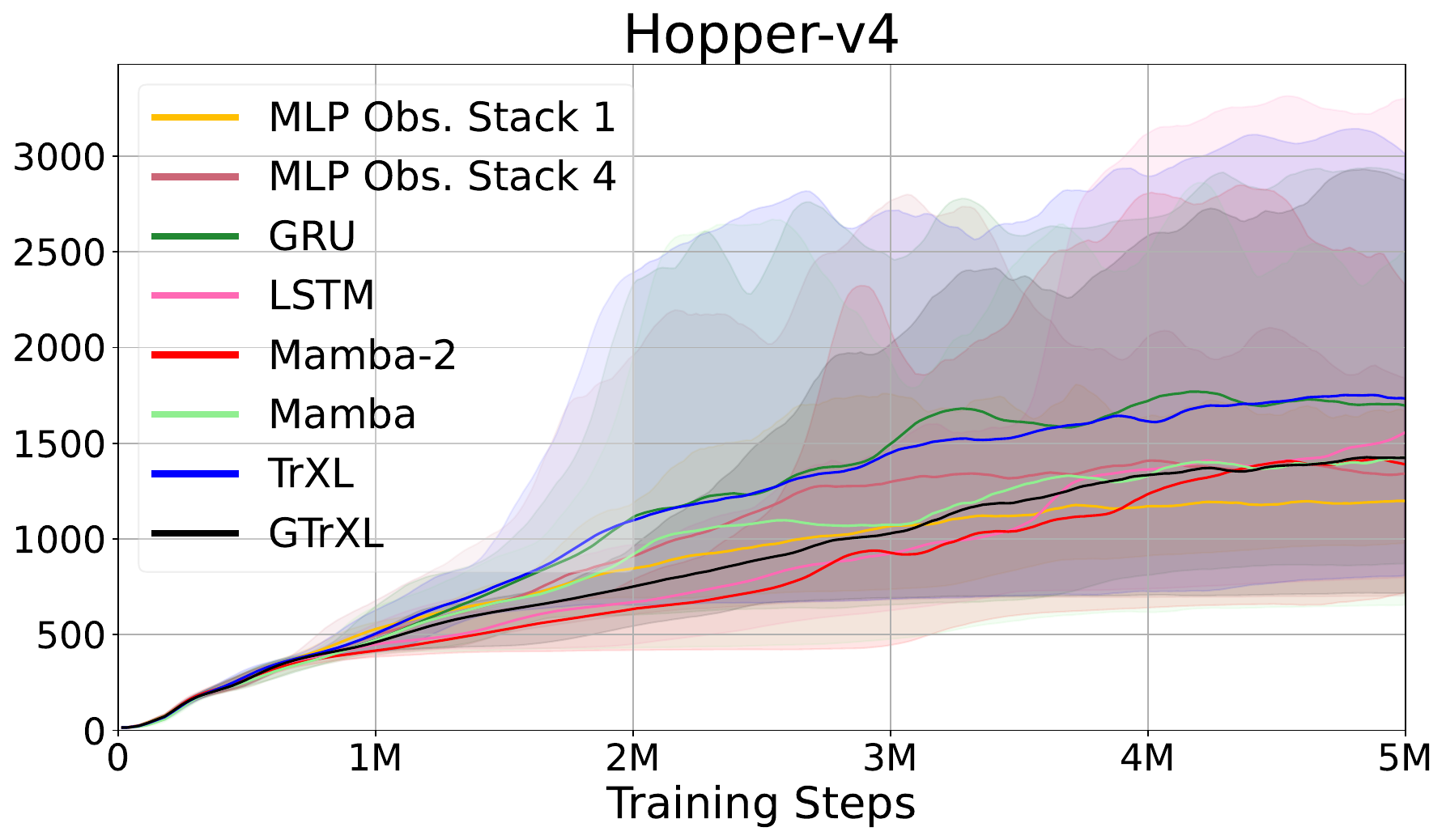}
\caption{Average returns for MuJoCo tasks. MLP and LSTM demonstrate competitive or superior performance in Walker2d and HalfCheetah, while GRU and Transformer-XL perform best in Hopper.}
\label{mujoco}
\end{figure}

Specifically, Figure~\ref{mujoco} presents learning curves for the MuJoCo environments (Walker2d-v4, HalfCheetah-v4, and Hopper-v4), revealing distinct architectural advantages across different continuous control tasks: (1) In \textbf{Walker2d-v4}, MLP achieves the best performance (approximately 3250), demonstrating outstanding stability and good sample efficiency. GRU and Mamba-2 follow closely, reflecting solid temporal modeling abilities, while Transformer-XL reaches a lower asymptote. Original Mamba trails substantially, suggesting optimization instability or inadequacy for complex motor control. (2) In \textbf{HalfCheetah-v4}, LSTM and PPO-1 consistently outperform all others, showing the highest reward (approximately 3350). The smooth progression indicates robustness and effective temporal integration in continuous action spaces. Transformer-XL, GRU, and Mamba-2 achieve moderate success but lag in both final reward and convergence speed, while the original Mamba significantly underperforms. (3) In \textbf{Hopper-v4}, GRU and Transformer-XL perform best, reaching high and stable rewards, showing that moderate complexity sequence models suit tasks requiring balance and stability. MLP performs worse than others, reinforcing the value of recurrence in tasks with strong stability constraints.

\subsection{Discrete Control Tasks}

 Mamba and Mamba-2 excel in discrete control tasks like Pong, achieving rapid convergence due to efficient state-space modeling. However, Mamba struggles in more complex scenarios like Breakout, where its limited temporal modeling becomes a disadvantage. LSTM and GRU demonstrate strong performance in strategic environments like Breakout, but their added complexity can slow learning in simpler tasks (e.g. Pong). Gated Transformer-XL (GTrXL) shows only minor improvements over Transformer-XL, indicating that the gating mechanism provides limited benefits in these tasks.

\begin{figure}[t]
\centering
\includegraphics[width=0.49\columnwidth]{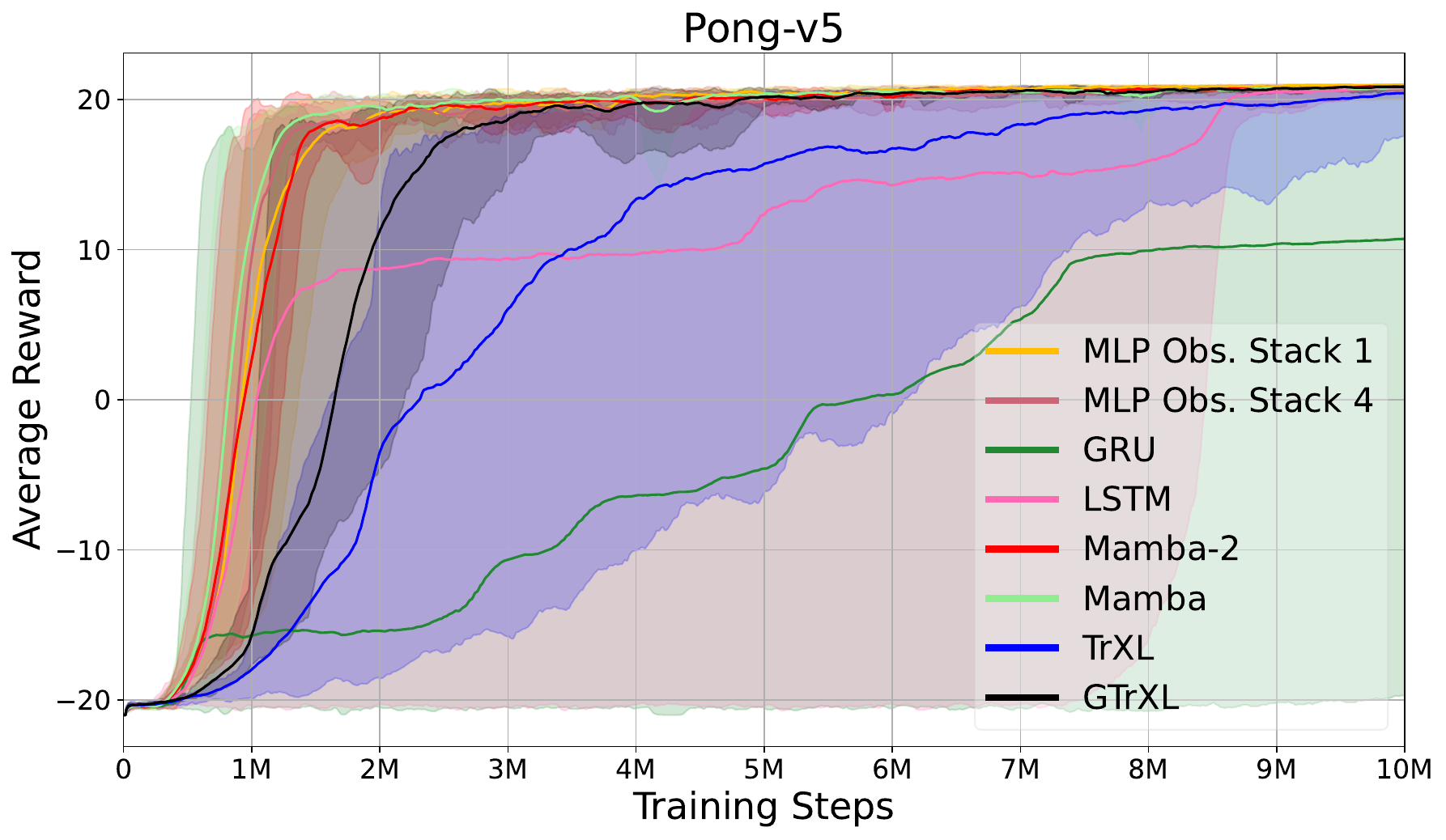}
\includegraphics[width=0.49\columnwidth]{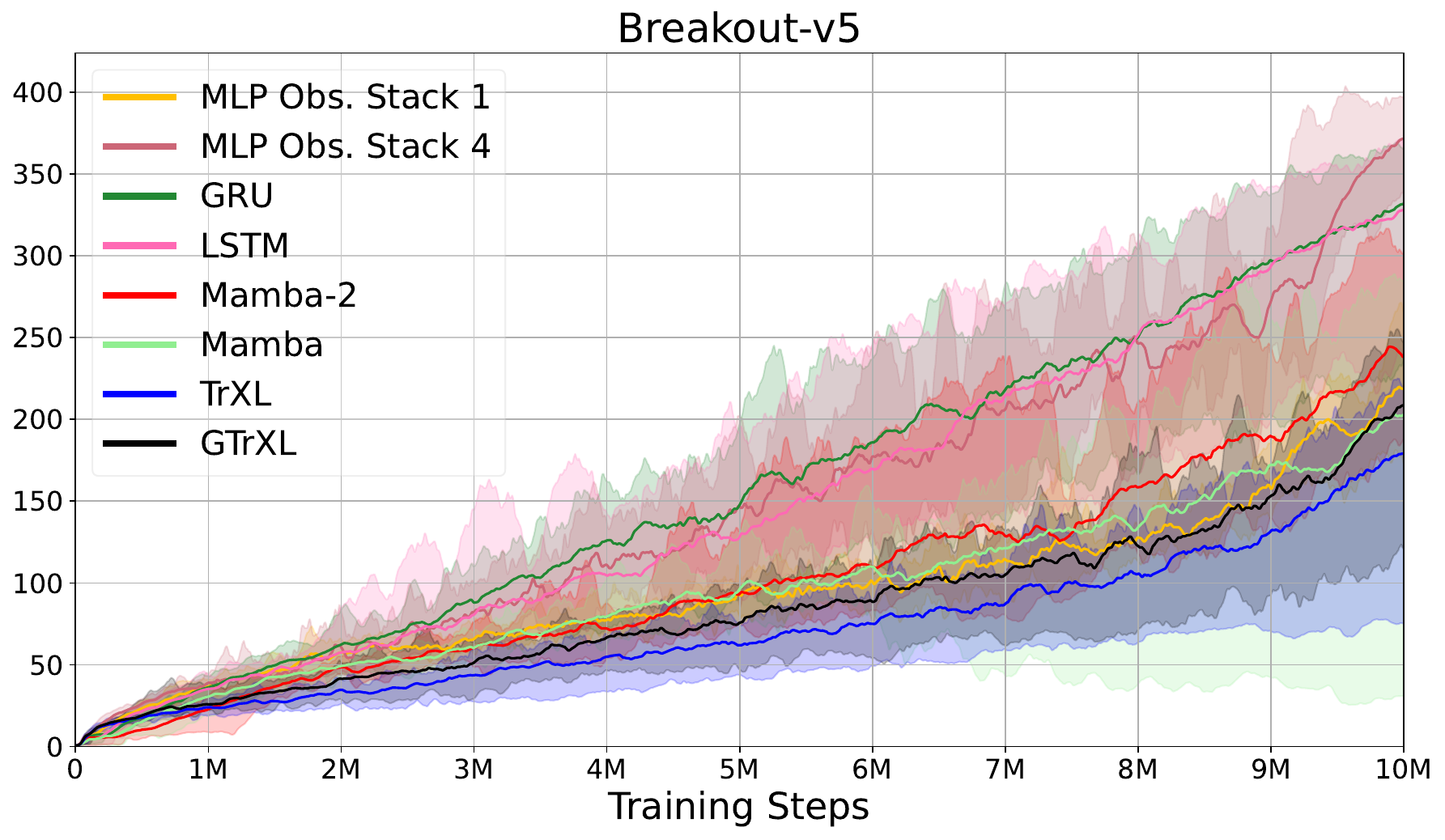}
\caption{Average returns across random seeds for Atari environments. Mamba and MLP with frame stacking excel in Pong, while LSTM and MLP with frame stacking perform best in Breakout.}
\label{atari}
\end{figure}

\begin{center}
\fcolorbox{black}{blue!10}{\parbox{.98\linewidth}{\textcolor{magenta}{Finding}: Mamba and Mamba-2 achieve fast convergence in reactive tasks like Pong but underperform in strategic environments like Breakout, where LSTM and GRU excel despite slower learning, while GTrXL offers only marginal gains over Transformer-XL.}}
\end{center}

For instance, results for Atari environments (Figure~\ref{atari}) reveal environment-specific architectural advantages: (1) In \textbf{Pong-v5}, Mamba, Mamba-2, and MLP rapidly reach maximum reward (20) with excellent sample efficiency (approximately 2M steps), indicating effectiveness in deterministic, reaction-time critical environments. GTrXL performs slightly better than standard Transformer-XL, suggesting that the gating mechanism enhances learning stability. LSTM and GRU exhibit slower learning, indicating that their recurrent nature introduces additional complexity in learning optimal behaviors, which may not be necessary for a straightforward, deterministic task like Pong. (2) For \textbf{Breakout-v5}, PPO-4 and recurrent architectures (GRU, LSTM) achieve the highest performance (approximately 360), steadily increasing reward and demonstrating good generalization and representation learning. All other architectures perform worse, implying difficulties with modeling task-specific structured temporal dynamics or input complexity in this Atari task.

\subsection{Partially Observable Control Tasks}

\begin{figure}[t]
\centering
\includegraphics[width=0.49\columnwidth]{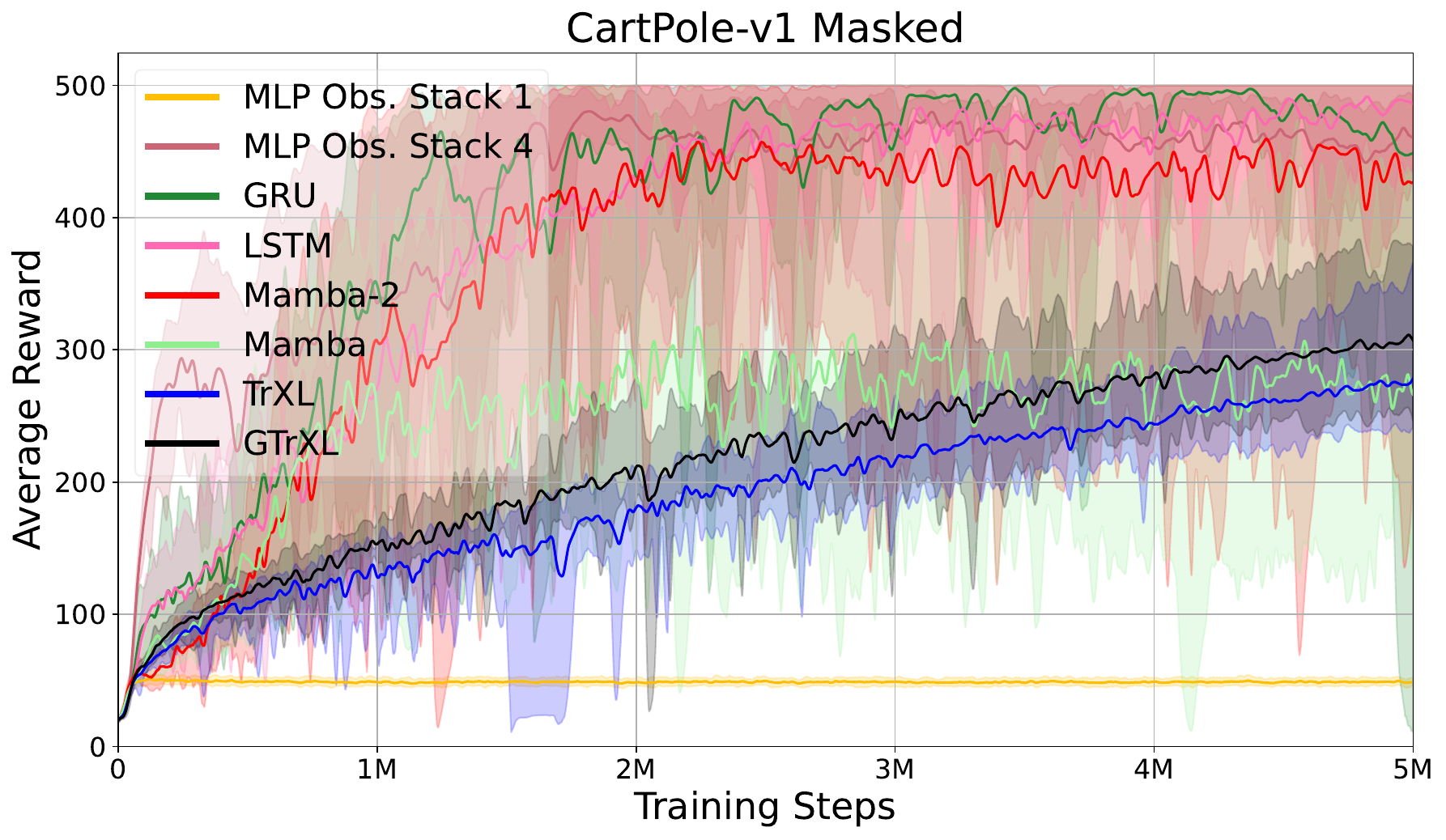}
\includegraphics[width=0.49\columnwidth]{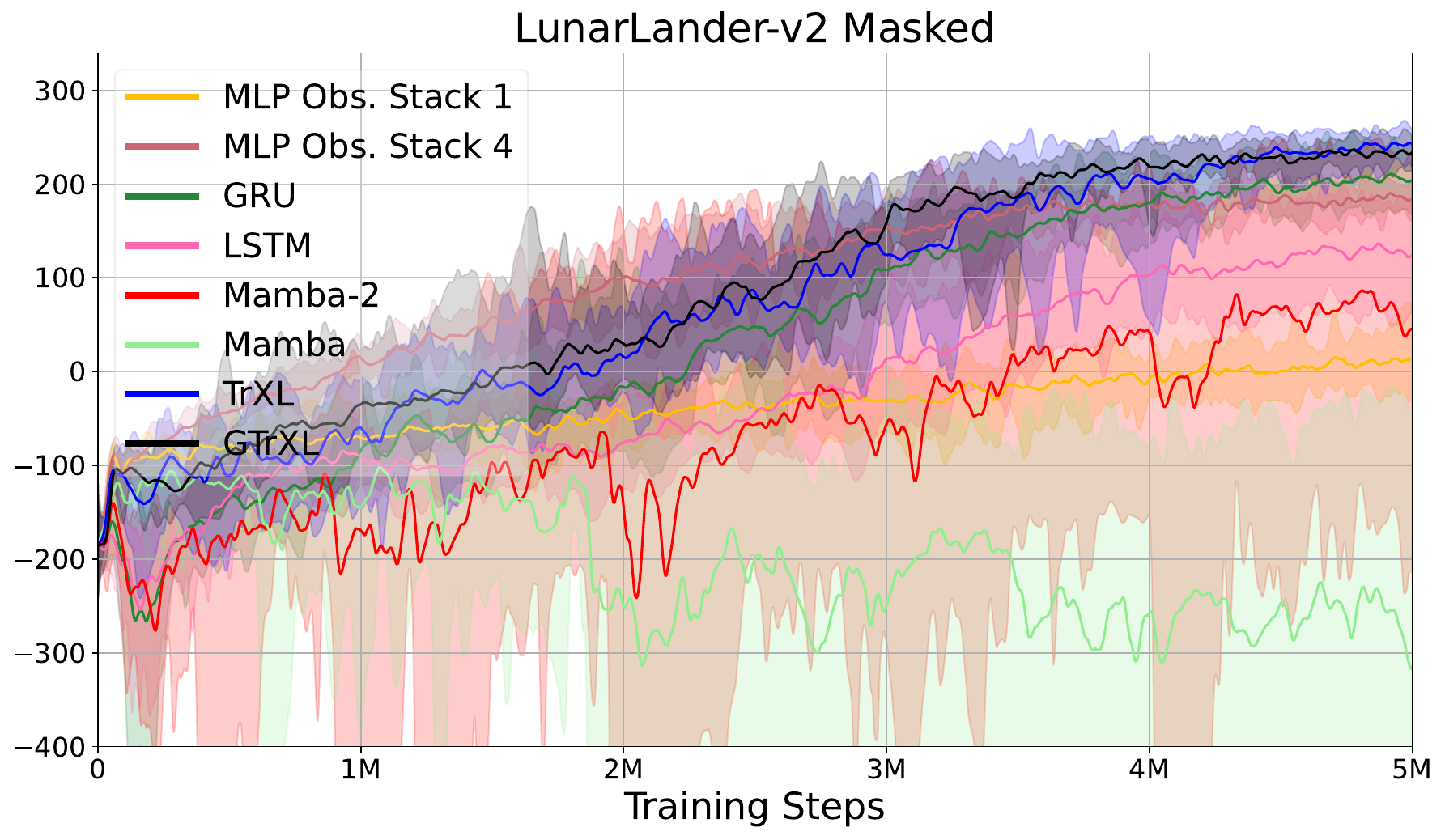}
\caption{Average returns across seeds for masked classic control tasks. Recurrent architectures and stacked MLPs excel in CartPole, while Transformer-XL performs best in LunarLander.}
\label{classic_control}
\end{figure}

Traditional recurrent architectures (GRU/LSTM) excel in simpler, short-horizon partially observable tasks. Transformer-XL demonstrates strong performance in complex, partially observable settings, effectively integrating information across longer time spans. GTrXL offers no substantial improvement over Transformer-XL, indicating limited benefits from the gating mechanism. Mamba exhibits poor performance and instability, potentially due to implementation challenges and information leakage across episodes. PPO-4 achieves fast and reliable performance, balancing simplicity and short-term memory. 

\begin{center}
\fcolorbox{black}{blue!10}{\parbox{.98\linewidth}{\textcolor{magenta}{Finding}: GRU and LSTM perform well in simple, short-horizon tasks, while Transformer-XL excels in complex, partially observable settings; GTrXL adds little benefit, Mamba struggles with instability, and PPO-4 offers fast, stable learning with minimal complexity.}}
\end{center}

For example, Figure~\ref{classic_control} shows results for masked classic control environments (CartPole-v1 and LunarLander-v2), where we removed all velocity information to create partial observability: (1) In \textbf{CartPole-v1 Masked}, GRU, LSTM, PPO-4, and Mamba-2 rapidly achieve maximum reward (approximately 500), demonstrating superior sample efficiency (approximately 1.5M steps) and robustness in masked, short-horizon tasks. (2) Transformer-XL has slower convergence and lower rewards, possibly due to overfitting or a lack of efficient representation of short-term masked inputs. (3) For \textbf{LunarLander-v2 Masked}, Transformer-XL/GTrXL, and PPO-4 are highly effective (200+ reward), especially Transformer-XL, which steadily overcomes partial observability (approximately 3M steps), revealing advantages of self-attention in inferring masked state elements. Mamba and Mamba-2 notably struggle, highlighting possible brittleness to masked or noisy observations. GRU and LSTM perform moderately well but plateau quickly, suggesting limitations in extracting masked information compared to self-attention mechanisms.

\subsection{Memory-Intensive Tasks}
\begin{figure}[t]
\centering
\includegraphics[width=0.49\columnwidth]{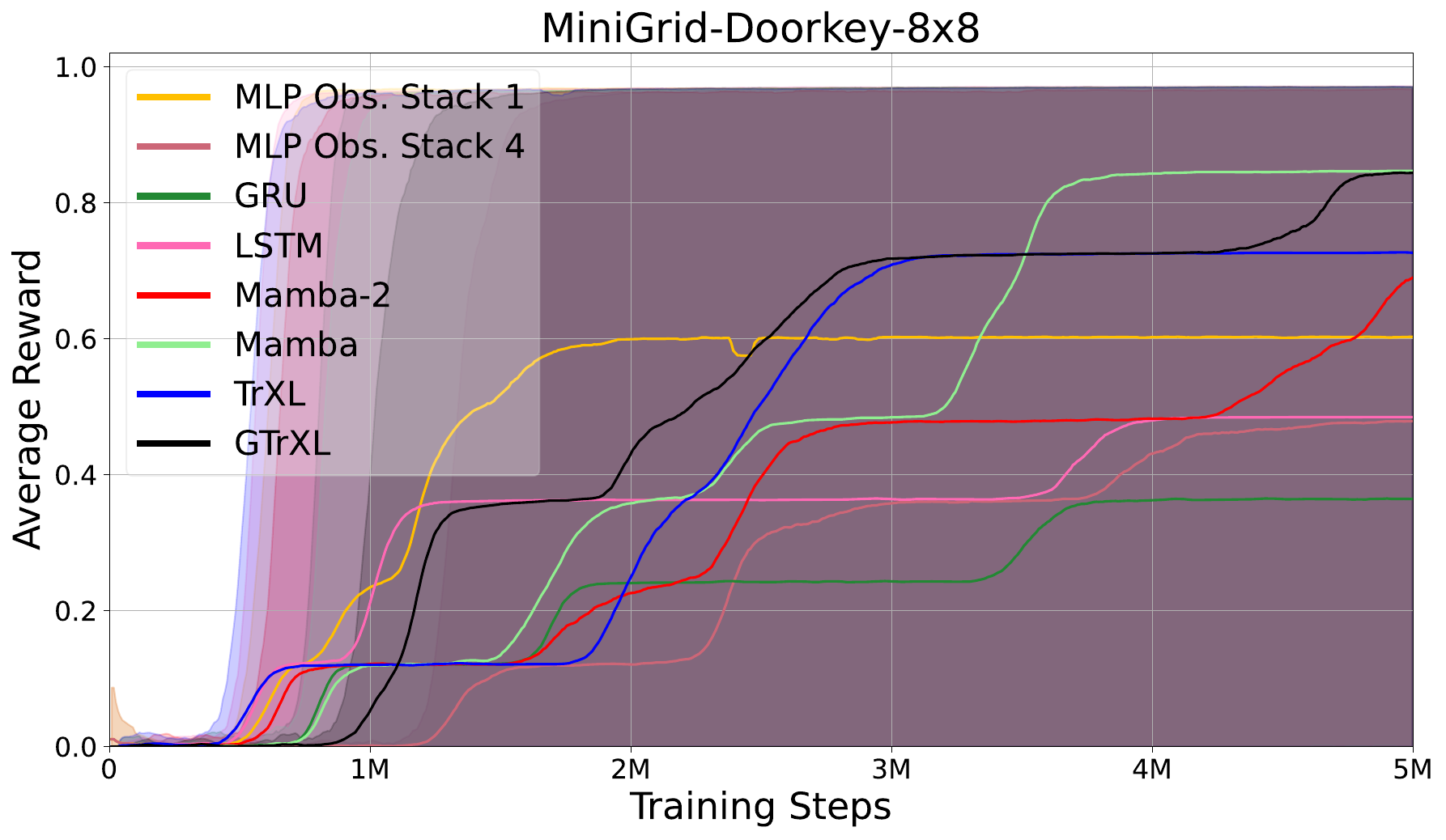}
\includegraphics[width=0.49\columnwidth]{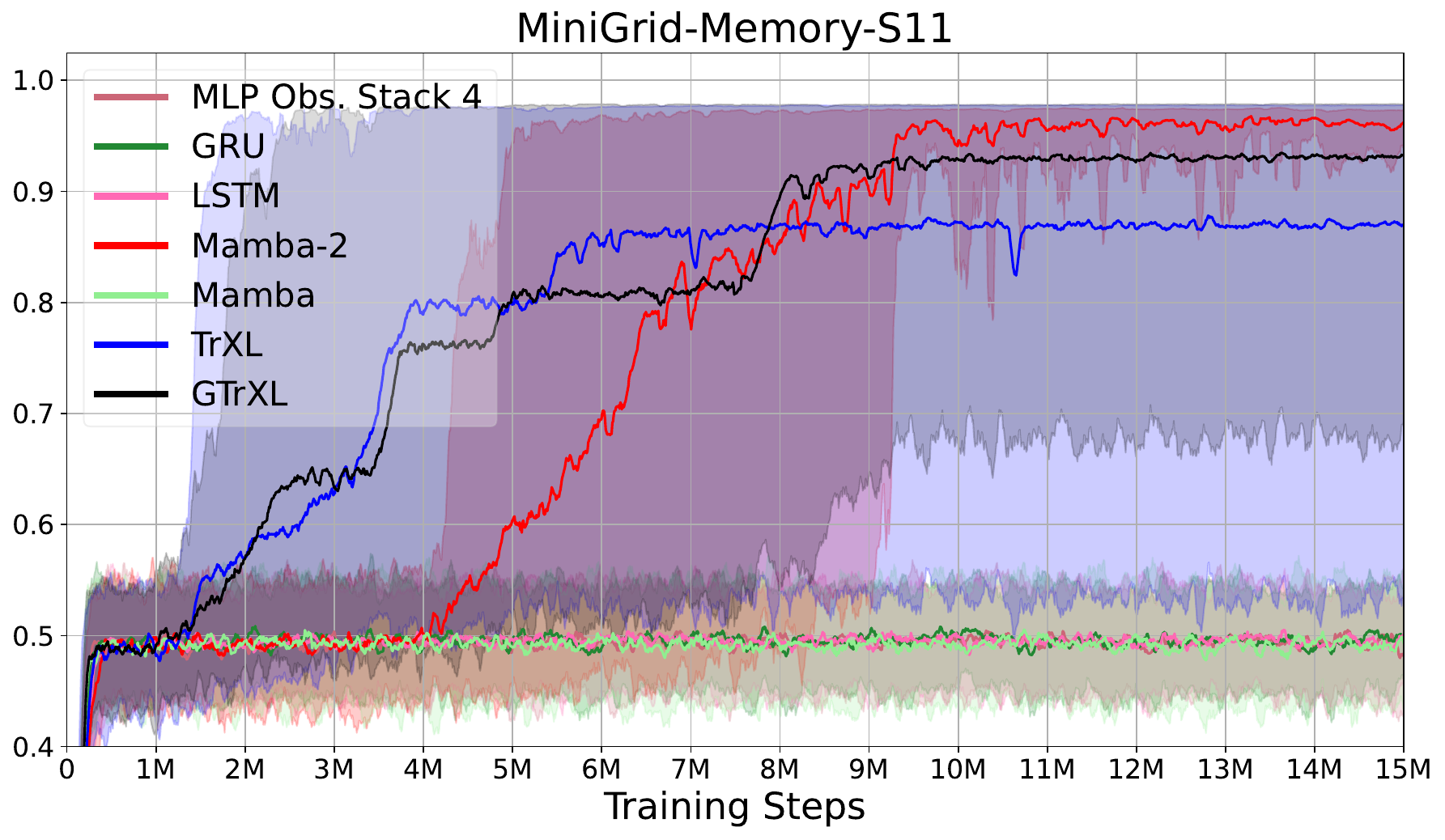}
\caption{Average returns across random seeds for MiniGrid environments. In DoorKey-8x8, original Mamba shows the fastest convergence, while in Memory-S11, only Transformer-XL and Mamba-2 achieve meaningful learning, with Mamba-2 reaching near-optimal performance.}
\label{minigrid}
\end{figure}

Based on the comprehensive experiment results, Mamba-2, Transformer-XL, and GTrXL are the only architectures capable of effectively solving long-horizon memory tasks, with GTrXL demonstrating more stable learning curves due to the gating mechanism. Both Mamba variants and Transformer-XL also excel in environments requiring complex credit assignment, such as DoorKey. In contrast, conventional recurrent architectures (LSTM, GRU) and simple MLP models struggle in these settings. For example, as shown in Figure~\ref{minigrid} presents results for our most memory-demanding environments: MiniGrid DoorKey-8x8 and Memory-S11. 

\begin{center}
\fcolorbox{black}{blue!10}{\parbox{.98\linewidth}{\textcolor{magenta}{Finding}: Mamba-2, Transformer-XL, and GTrXL are uniquely effective in long-horizon memory and complex credit assignment tasks, with GTrXL offering added stability via gating, while traditional recurrent and MLP models fall short. }}
\end{center}

In \textbf{DoorKey-8x8}, the original Mamba architecture demonstrates remarkable sample efficiency, rapidly converging to the highest reward. Its learning curve sharply surpasses other models, indicating superior mid-length memory and efficient representation learning. Transformer-XL and GTrXL achieve a strong result slightly slower, demonstrating good generalizability but moderate sample efficiency. Mamba-2 achieves moderate success more slowly, suggesting potential optimization challenges on shorter-horizon memory tasks compared to original Mamba. Surprisingly, a simple PPO-1 outperforms standard recurrent architectures (LSTM, GRU), suggesting that minimal state abstraction is sufficient in less complex tasks.

In the highly challenging \textbf{Memory-S11} environment, Mamba-2 significantly outperforms all other architectures, achieving near-optimal reward (approximately 0.96) with remarkable stability. Transformer-XL exhibits steady and stable performance but converges at a slightly lower optimal reward with higher variance. GTrXL achieves similar results but get higher mean reward, indicating that gating mechanism provides more stable learning curves. LSTM, GRU, MLP, and original Mamba show no meaningful learning beyond random exploration, indicating their limited capabilities in tasks requiring extensive memory.

\subsection{Computational Efficiency Analysis}
\label{sec}
The detailed measurements of computational efficiency across architectures are presented in the Appendix. These metrics are critical for evaluating the practical applicability of each approach in resource-constrained settings:

\begin{center}
\fcolorbox{black}{blue!10}{\parbox{.98\linewidth}{\textcolor{magenta}{Finding}: Mamba models are ideal for resource-constrained environments where fast throughput and low memory usage are critical compared with other architectures such as Transformer XL, LSTM and GRU.}}
\end{center}

Mamba achieves exceptional computational efficiency, being 4.5× faster than LSTM, 3.9× faster than GRU, and 1.5× faster than Transformer-XL, while maintaining low memory usage (8× less than Transformer-XL). Mamba-2, despite being slightly slower, retains significant efficiency advantages over LSTM and Transformer-XL.

Mamba achieves an average of 2734 \textit{steps per second (SPS)}, which is significantly faster than LSTM, GRU, and Transformer-XL, though still slower than MLP (1.3×). This throughput advantage translates directly to training time improvements, with Mamba completing the same number of environment interactions in approximately one-quarter the time required by LSTM.

In terms of \textit{inference latency} (reported in the Appendix), Mamba maintains relatively low response times (1.30 ms on average), comparable to recurrent models such as LSTM (1.01 ms) and GRU (0.971 ms). While Mamba is approximately 1.3× slower than GRU and LSTM, it is still significantly faster (1.66×) than Transformer-XL, which averages 2.17 ms. Despite these differences, all models demonstrate low-latency performance overall, and such variations are unlikely to pose significant challenges in most real-world applications where fast decision-making is required.

\textit{Memory efficiency} shows perhaps the most dramatic differences between architectures. Mamba requires only 0.217 GB of GPU memory on average, which is 8.1× less than Transformer-XL (1.765 GB) while achieving comparable or superior performance in most environments. This substantial memory advantage makes Mamba suitable for deployment on resource-constrained edge devices or for scaling to larger batch sizes on standard hardware.

\subsection{Architecture-Environment Compatibility}
Our comprehensive evaluation reveals clear patterns regarding which architectures excel in particular environments:

\textbf{Memory-independent tasks:} In environments with relatively smooth or Markovian dynamics, such as continuous control tasks (Walker2d, HalfCheetah) and reaction-based games (Pong), simpler architectures like MLPs and Mamba perform effectively by capturing immediate dependencies with high stability. However, for tasks requiring strategic planning (e.g., Breakout), models with recurrent structure or stacked inputs (like LSTM and MLP Stack-4) are better suited.

\textbf{Partially observable environments:} The optimal architecture depends on the complexity of the hidden state. In simpler masked tasks (CartPole), traditional recurrent architectures (GRU/LSTM) excel, while more complex partially observable environments (LunarLander) benefit from attention mechanisms (Transformer-XL) that can more effectively infer hidden variables.

\textbf{Memory tasks:} In environments requiring moderate memory capabilities (DoorKey-8x8), the original Mamba architecture demonstrates outstanding sample efficiency and performance, suggesting its selective state-space approach provides an ideal inductive bias for mid-length memory requirements. However in long-horizon tasks only Mamba-2 and Transformer-XL can effectively solve tasks requiring extensive memory (Memory-S11), with Mamba-2 achieving superior performance. This indicates that advanced state-space models with selective attention mechanisms are uniquely suited to long-term dependency modeling in RL.

These patterns provide actionable guidance for practitioners: architecture selection should be driven by the specific memory and control requirements of the target environment, with simpler architectures preferred unless the task specifically demands long-term memory retention or complex partially observable state inference.

\subsection{Practical Guidelines for Practitioners}
Based on our comprehensive evaluation, we propose the following guidelines for selecting neural architectures in reinforcement learning:

(1) \textbf{Start with MLP}: For most tasks, particularly those with largely Markovian dynamics (e.g., MuJoCo), Multi-Layer Perceptrons (MLPs) provide an excellent balance of performance, stability, and computational efficiency. They are fast to train and offer strong baseline performance.

(2) \textbf{Prioritize Mamba-2 for Sequence Tasks}: If the task involves temporal dependencies or partial observability, Mamba-2 should be your first choice. It offers a unique combination of fast training (approximately 5× faster than LSTM/GRU) and competitive performance, making it a practical first option for sequence modeling.

(3) \textbf{Explore LSTM and GRU if Mamba-2 Falls Short}: In cases where Mamba-2 does not achieve satisfactory results, consider trying LSTM and GRU. While they require significantly longer training times, they may outperform Mamba-2 in some environments due to their well-established memory modeling capabilities.

(4) \textbf{Reserve Transformers for Challenging Memory Tasks}: Transformers (Transformer-XL, GTrXL) should be considered only for environments that are extremely memory-intensive, such as long-horizon planning or complex partially observable tasks. Their high computational cost and implementation complexity make them unsuitable for most practical applications unless the task is specifically designed to benefit from long-range memory modeling.

These guidelines serve as a starting point, but optimal architecture selection should ultimately depend on the specific characteristics of the target environment and the available computational resources.

\subsection{Future Work}

This work opens several directions for further research. First, a more thorough hyperparameter optimization process — beyond default CleanRL settings — could provide a fairer comparison across architectures, particularly for Transformer-XL and Mamba, which may benefit from task-specific tuning. Second, architectural ablation studies are needed to isolate the contributions of specific components, such as recurrence depth, attention heads, or state-space scan mechanisms, to better understand performance–efficiency tradeoffs. 

Another important technical improvement involves adding proper hidden state resets to Mamba, as our current implementation allows information leakage between episodes, potentially affecting performance in episodic tasks. Future work should also explore deeper or stacked versions of each architecture to investigate scaling behaviors. Finally, extending our evaluation beyond PPO to other RL algorithms (e.g., SAC \cite{haarnoja2018softactorcriticoffpolicymaximum}, TD3 \cite{fujimoto2018td3}) and exploring hybrid architectures that combine recurrence, attention, and state-space memory could lead to more flexible and robust solutions for partially observable environments.

\section{Conclusion}
Our systematic evaluation of neural architectures in reinforcement learning reveals fundamental patterns with direct implications for real-world applications:

(1) Mamba, despite its efficient design, shows inconsistent performance across environments in our experiments. However, Mamba-2 consistently delivers strong results, combining high performance in memory-intensive tasks with exceptional computational efficiency. It achieves competitive performance with LSTM and GRU while offering substantially faster training (5.3× faster) and lower memory requirements (8.1× less GPU memory than Transformer-XL).

(2) Simpler architectures like MLPs remain highly competitive in Markovian environments (e.g., MuJoCo), where their low complexity and fast training are advantageous. For tasks without clear temporal dependencies, starting with an MLP is both practical and efficient.

(3) When some temporal dependencies are expected, Mamba-2 is a strong initial choice due to its efficient state-space design and fast training. If Mamba-2 shows instability or underperformance, LSTM and GRU can be explored next, offering robust recurrent modeling capabilities.

(4) Transformers (e.g., Transformer-XL, GTrXL) are both more complex to set up and more memory-intensive, making them a suitable choice only for environments with extensive memory requirements (e.g., MiniGrid Memory-S11). Among these, GTrXL is generally preferred — it delivers slightly better final performance, smoother learning curves, and lower variance.

(5) Most importantly, our findings challenge the assumption that increased architectural complexity translates to superior performance. The strong results of MLPs and Mamba-2 suggest that simpler or more efficient architectures are often preferable, especially in resource-constrained settings. Practitioners are encouraged to start with efficient models and only scale complexity when the task demands it.

\bibliography{sources/references}
\bibliographystyle{plain}


\newpage
\appendix
\onecolumn
\section{Performance Metrics.}

\begin{table}[h!]
\centering
\small
\caption{Final Steps Per Second (SPS) for Various Architectures and Environments (Rounded)}
\begin{adjustbox}{width=1.\textwidth,center}
\begin{tabular}{lcccccccc}
\toprule
Environment & PPO-1 & PPO-4 & LSTM & GRU & TrXL & GTrXL & Mamba & Mamba-2 \\
\midrule
MiniGrid-MemoryS11-v0   & 3191 & 2544 & 802 & 924 & 1697 & 1869 & 2850 & 2698 \\
MiniGrid-DoorKey-8x8-v0 & 3496 & 2770 & 828 & 957 & 1684 & 1868 & 3031 & 2844 \\
Breakout-v5             & 1521 & 1487 & 626 & 701 & 1188 & 1179 & 1329 & 1245 \\
Pong-v5                 & 1753 & 1699 & 666 & 742 & 1322 & 1310 & 1512 & 1404 \\
CartPole-v1             & 7046 & 6930 & 991 & 1151 & 3781 & 3594 & 4643 & 3827 \\
LunarLander-v2          & 5979 & 5726 & 897 & 1061 & 2585 & 2489 & 4018 & 3398 \\
Walker2d-v4             & 2738 & 2675 & 210 & 256 & 1430 & 1436 & 2270 & 2121 \\
HalfCheetah-v4          & 3315 & 3212 & 211  & 259  & 1577 & 1706 & 2632 & 2429 \\
Hopper-v4               & 2808 & 2703 & 208  & 256  & 1440 & 1555 & 2325 & 2133 \\
\midrule
\textbf{Average}        & \textbf{3539} & \underline{3305} & 604 & 701 & 1856 & 1890 & 2734 & 2455 \\
\bottomrule
\end{tabular}
\end{adjustbox}
\label{tab:env_sps_final}
\end{table}

\begin{table}[h!]
\centering
\caption{Training Time (in Minutes) for Various Architectures and Environments (3 million total timesteps)}
\begin{adjustbox}{width=0.93\textwidth,center}
\begin{tabular}{lcccccccc}
\toprule
Environment & PPO-1 & PPO-4 & LSTM & GRU & TrXL & GTrXL & Mamba & Mamba-2 \\
\midrule
MiniGrid-DoorKey-8x8-v0 & 14.31 & 18.06 & 60.33 & 52.22 & 29.68 & 26.76 & 16.50 & 17.59 \\
MiniGrid-MemoryS11-v0   & 15.67 & 19.66 & 62.31 & 54.04 & 29.46 & 26.74 & 17.75 & 18.54 \\
Breakout-v5             & 32.91 & 33.65 & 79.78 & 71.29 & 42.10 & 42.43 & 37.66 & 40.17 \\
Pong-v5                 & 28.55 & 29.46 & 75.03 & 67.34 & 37.83 & 38.20 & 33.10 & 35.64 \\
CartPole-v1             & 7.11 & 7.23 & 50.41 & 43.43 & 13.23 & 13.92 & 10.78 & 13.07 \\
LunarLander-v2          & 8.37 & 8.74 & 55.66 & 47.11 & 19.34 & 20.10 & 12.45 & 14.72 \\
Walker2d-v4             & 18.27 & 18.70   & 237.78 & 194.69 & 34.95 & 35.13 & 22.03 & 23.58 \\
HalfCheetah-v4          & 15.10 & 15.58   & 236.42  & 192.39  & 31.70 & 29.32 & 19.00 & 20.59 \\
Hopper-v4          & 17.82 & 18.51   & 239.37  & 96.81  & 34.70 & 32.15 & 21.51 & 23.44 \\
\midrule
\textbf{Average}        & \textbf{16.59} & \underline{18.84} & 121.90 & 91.04 & 30.33 & 29.42 & 21.20 & 22.97 \\
\bottomrule
\end{tabular}
\end{adjustbox}
\label{tab:env_training_time_minutes}
\end{table}

\begin{table}[h!]
\centering
\caption{Evaluation Results (Mean $\pm$ Std) of Final Average Episode Return.}
\begin{adjustbox}{width=1.\textwidth,center}
\begin{tabular}{lcccccccc}
\toprule
\textbf{Environment} & PPO-1 & PPO-4 & LSTM & GRU & TrXL & GTrXL & Mamba & Mamba-2 \\
\midrule
\multicolumn{8}{l}{\textbf{MiniGrid}} \\
MemoryS11   & -- & 0.49 $\pm$ 0.02 & 0.51 $\pm$ 0.04 & 0.49 $\pm$ 0.02 & 0.88 $\pm$ 0.18 & \underline{0.93 $\pm$ 0.11} & 0.49 $\pm$ 0.03 & \textbf{0.96 $\pm$ 0.01} \\
DoorKey     & 0.60 $\pm$ 0.50 & 0.48 $\pm$ 0.51 & 0.49 $\pm$ 0.52 & 0.36 $\pm$ 0.50 & 0.73 $\pm$ 0.45 & \underline{0.84 $\pm$ 0.34} & \textbf{0.85 $\pm$ 0.34} & 0.69 $\pm$ 0.43 \\
\midrule
\multicolumn{8}{l}{\textbf{Atari}} \\
Breakout    & 220.5 $\pm$ 27.6 & \textbf{372.7 $\pm$ 20.3} & 327.8 $\pm$ 38.4 & \underline{332.2 $\pm$ 43.4} & 180.4 $\pm$ 55.5 & 208.7 $\pm$ 38.3 & 202.9 $\pm$ 89.0 & 239.2 $\pm$ 39.5 \\
Pong        & \textbf{20.98 $\pm$ 0.03} & 20.69 $\pm$ 0.33 & 20.75 $\pm$ 0.14 & 10.71 $\pm$ 18.76 & 20.41 $\pm$ 1.20 & 20.81 $\pm$ 0.1 & 20.82 $\pm$ 0.10 & \underline{20.89 $\pm$ 0.09} \\
\midrule
\multicolumn{8}{l}{\textbf{Classic Control}} \\
Cartpole    & 49.4 $\pm$ 2.1 & \underline{458.7 $\pm$ 28.9} & \textbf{484.6 $\pm$ 26.6} & 449.4 $\pm$ 152.8 & 281.8 $\pm$ 41.2 & 305.8 $\pm$ 40.6 & 272.8 $\pm$ 104.7 & 434.2 $\pm$ 90.5 \\
LunarLander & 16.6 $\pm$ 28.9 & 190.1 $\pm$ 11.8 & 123.8 $\pm$ 38.0 & 204.8 $\pm$ 23.2 & \textbf{244.9 $\pm$ 13.1} & \underline{236.5 $\pm$ 8.5} & -313.2 $\pm$ 235.1 & 42.5 $\pm$ 182.2 \\
\midrule
\multicolumn{8}{l}{\textbf{MuJoCo}} \\
HalfCheetah     & \underline{3918.8 $\pm$ 312.8} & 3116.4 $\pm$ 612.0 & \textbf{3997.4 $\pm$ 1378.8} & 2745.0 $\pm$ 1118.1 & 3240.8 $\pm$ 848.8 & 3464.0 $\pm$ 921.7 & 2238.7 $\pm$ 627.2 & 3718.0 $\pm$ 528.5 \\
Hopper      & 1194.1 $\pm$ 200.4 & 1366.4 $\pm$ 495.1 & 1574.4 $\pm$ 842.7 & \underline{1676.6 $\pm$ 724.5} & \textbf{1712.0 $\pm$ 741.6} & 1428.5 $\pm$ 711.4 & 1409.2 $\pm$ 635.7 & 1390.2 $\pm$ 639.3 \\
Walker2d    & \textbf{3379.3 $\pm$ 1039.8} & 3152.1 $\pm$ 193.0 & 3056.1 $\pm$ 547.4 & 3170.3 $\pm$ 355.4 & \underline{3206.1 $\pm$ 528.0} & 2765.7 $\pm$ 937.4 & 2512.6 $\pm$ 1000.5 & 3038.0 $\pm$ 505.2 \\
\bottomrule
\end{tabular}
\end{adjustbox}
\label{tab:episode_return_results}
\end{table}

\begin{table}[h!]
\centering
\caption{Number of Parameters (in Thousands) for Various Architectures and Environments}
\begin{adjustbox}{width=1.\textwidth,center}
\begin{tabular}{lccccccc}
\toprule
Environment & PPO-1 & LSTM & GRU & TrXL & GTrXL & Mamba & Mamba-2 \\
\midrule
Hopper-v4               & 39.7 & 38.6 & 39.1 & 37.8 & 44.9 & 40.1 & 43.8 \\
HalfCheetah-v4          & 40.4 & 39.2 & 39.7 & 38.3 & 45.5 & 40.7 & 44.4 \\
Walker2d-v4             & 40.4 & 39.2 & 39.7 & 38.3 & 45.5 & 40.7 & 44.4 \\
Pong-v5                 & 2527.2 & 2468.8 & 2271.7 & 2669.2 & 2639.3 & 2413.9 & 2805.3 \\
Breakout-v5             & 2731.2 & 2468.3 & 2271.1 & 2668.2 & 2638.4 & 2413.0 & 2804.3 \\
LunarLander-v2          & 1057.8 & 1057.8 & 1042.8 & 1035.3 & 1070.1 & 1089.0 & 973.0 \\
CartPole-v1             & 265.2 & 265.5 & 268.6 & 261.9 & 264.5 & 262.6 & 226.0 \\
MiniGrid-MemoryS11-v0   & 2470.9 & 2473.1 & 2276.0 & 2408.4 & 2591.7 & 2470.9 & 2426.0 \\
MiniGrid-DoorKey-8x8-v0 & 2531.8 & 2473.1 & 2465.2 & 2408.4 & 2510.6 & 2357.8 & 2426.0 \\
\bottomrule
\end{tabular}
\end{adjustbox}
\label{tab:env_params_k}
\end{table}

\begin{table}[h!]
\centering
\caption{Inference Latency (ms) for Various Architectures and Environments}
\begin{adjustbox}{width=1.\textwidth,center}
\begin{tabular}{lcccccccc}
\toprule
Environment & PPO-1 & PPO-4 & LSTM & GRU & TrXL & GTrXL & Mamba & Mamba-2 \\
\midrule
MiniGrid-MemoryS11-v0   & 1.139 & 1.309 & 1.290 & 1.254 & 2.917 & 2.729 & 1.603 & 1.780 \\
MiniGrid-DoorKey-8x8-v0 & 1.093 & 1.249 & 1.215 & 1.171 & 2.845 & 2.666 & 1.534 & 1.754 \\
Breakout-v5             & 1.016 & 1.040 & 1.180 & 1.127 & 1.792 & 1.941 & 1.495 & 1.668 \\
Pong-v5                 & 1.024 & 1.047 & 1.167 & 1.132 & 1.775 & 1.946 & 1.488 & 1.690 \\
CartPole-v1             & 0.739 & 0.729 & 0.909 & 0.887 & 1.455 & 1.600 & 1.158 & 1.347 \\
LunarLander-v2          & 0.725 & 0.736 & 0.956 & 0.920 & 1.973 & 2.249 & 1.176 & 1.367 \\
Walker2d-v4             & 0.658 & 0.660 & 0.776 & 0.748 & 2.268 & 2.073 & 1.093 & 1.260 \\
HalfCheetah-v4          & 0.649 & 0.656 & 0.780 & 0.747 & 2.245 & 2.056 & 1.099 & 1.263 \\
Hopper-v4               & 0.659 & 0.666 & 0.780 & 0.749 & 2.272 & 2.061 & 1.093 & 1.272 \\
\midrule
\textbf{Average}        & \textbf{0.856} & \underline{0.899} & 1.006 & 0.971 & 2.171 & 2.147 & 1.304 & 1.489 \\
\bottomrule
\end{tabular}
\end{adjustbox}
\label{tab:inference_latency}
\end{table}

\begin{table}[h!]
\centering
\caption{GPU Memory Allocated (GB) for Various Architectures and Environments}
\begin{adjustbox}{width=1.\textwidth,center}
\begin{tabular}{lcccccccc}
\toprule
Environment & PPO-1 & PPO-4 & LSTM & GRU & TrXL & GTrXL & Mamba & Mamba-2 \\
\midrule
MiniGrid-MemoryS11-v0   & 0.702 & 2.644 & 0.705 & 0.701 & 5.729 & 4.371 & 0.788 & 0.799 \\
MiniGrid-DoorKey-8x8-v0 & 0.702 & 2.644 & 0.705 & 0.705 & 2.341 & 1.870 & 0.796 & 0.799 \\
Pong-v5                 & 0.113 & 0.274 & 0.116 & 0.115 & 0.712 & 0.636 & 0.125 & 0.125 \\
Breakout-v5             & 0.113 & 0.274 & 0.116 & 0.116 & 0.730 & 0.650 & 0.125 & 0.125 \\
CartPole-v1             & 0.020 & 0.020 & 0.021 & 0.021 & 0.353 & 0.282 & 0.023 & 0.025 \\
LunarLander-v2          & 0.020 & 0.020 & 0.036 & 0.035 & 1.034 & 0.804 & 0.039 & 0.043 \\
Walker2d-v4             & 0.018 & 0.022 & 0.018 & 0.018 & 1.657 & 1.125 & 0.020 & 0.020 \\
HalfCheetah-v4          & 0.018 & 0.022 & 0.018 & 0.018 & 1.648 & 1.108 & 0.020 & 0.020 \\
Hopper-v4               & 0.018 & 0.021 & 0.018 & 0.018 & 1.677 & 1.128 & 0.019 & 0.019 \\
\midrule
\textbf{Average}        & \textbf{0.035} & 0.660 & \underline{0.194} & \underline{0.194} & 1.765 & 1.330 & 0.217 & 0.219 \\
\bottomrule
\end{tabular}
\end{adjustbox}
\label{tab:gpu_memory_allocated}
\end{table}

\begin{table}[h!]
\centering
\caption{GPU Memory Reserved (GB) for Various Architectures and Environments}
\begin{adjustbox}{width=1.\textwidth,center}
\begin{tabular}{lcccccccc}
\toprule
Environment & PPO-1 & PPO-4 & LSTM & GRU & TrXL & GTrXL & Mamba & Mamba-2 \\
\midrule
MiniGrid-MemoryS11-v0   & 1.133 & 3.941 & 1.131 & 1.203 & 18.630 & 15.707 & 1.236 & 1.285 \\
MiniGrid-DoorKey-8x8-v0 & 1.133 & 3.941 & 1.131 & 1.127 & 8.494  & 12.556 & 1.238 & 1.285 \\
Pong-v5                 & 0.264 & 0.393 & 0.273 & 0.268 & 2.002  & 1.781 & 0.285 & 0.396 \\
Breakout-v5             & 0.264 & 0.391 & 0.273 & 0.268 & 2.204  & 1.984 & 0.283 & 0.396 \\
CartPole-v1             & 0.027 & 0.027 & 0.049 & 0.051 & 1.089  & 0.970 & 0.051 & 0.512 \\
LunarLander-v2          & 0.027 & 0.027 & 0.068 & 0.066 & 2.955  & 2.292 & 0.063 & 0.541 \\
Walker2d-v4             & 0.031 & 0.051 & 0.053 & 0.053 & 5.461  & 3.631 & 0.033 & 0.514 \\
HalfCheetah-v4          & 0.031 & 0.051 & 0.053 & 0.053 & 3.668  & 2.504 & 0.033 & 0.514 \\
Hopper-v4               & 0.031 & 0.029 & 0.053 & 0.053 & 5.070  & 3.288 & 0.033 & 0.514 \\
\midrule
\textbf{Average}        & \textbf{0.327} & 0.983 & \underline{0.343} & 0.349 & 5.508 & 4.968 & 0.362 & 0.662 \\
\bottomrule
\end{tabular}
\end{adjustbox}
\label{tab:gpu_memory_reserved}
\end{table}

\section{Hyperparameters}
\label{appendix:hyperparams}

For reproducibility and transparency, we list the detailed hyperparameters and training settings used throughout our experimental evaluation:

\begin{table}[h]
    \centering
    \renewcommand{\arraystretch}{1.5} 
    \caption{Training settings. Steps and seeds to the right are for computational metrics.}
    \label{tab:env_settings}
    \begin{adjustbox}{width=0.65\textwidth,center}
    \begin{tabular}{llrr|rr}
        \hline
        \textbf{Domain} & \textbf{Environment} & \textbf{Steps} & \textbf{Seeds} & \textit{Steps} & \textit{Seeds} \\
        \hline
        Classic Control & CartPole-v1 & 5M & 10 & \textit{3M} & \textit{3} \\
        Classic Control & LunarLander-v2 & 5M & 10 & \textit{3M} & \textit{3} \\
        \hline
        Atari & Pong-v5 & 10M & 8 & \textit{3M} & \textit{3} \\
        Atari & Breakout-v5 & 10M & 8 & \textit{3M} & \textit{3} \\
        \hline
        Mujoco & HalfCheetah-v4 & 5M & 8 & \textit{3M} & \textit{3} \\
        Mujoco & Hopper-v4 & 5M & 8 & \textit{3M} & \textit{3} \\
        Mujoco & Walker2d-v4 & 5M & 8 & \textit{3M} & \textit{3} \\
        \hline
        MiniGrid & MemoryS11-v0 & 15M & 8 & \textit{3M} & \textit{3} \\
        MiniGrid & DoorKey-8x8-v0 & 5M & 8 & \textit{3M} & \textit{3} \\
        \hline
    \end{tabular}
    \end{adjustbox}
\end{table}

\begin{table}[h!]
\centering
\small
\caption{Common Training Hyperparameters by Domain}
\label{tab:common_hyperparams}
\begin{tabular}{lcccc}
\toprule
\textbf{Parameter}                   & \textbf{MiniGrid}         & \textbf{MuJoCo}           & \textbf{Classic Control}   & \textbf{Atari}             \\
\midrule
Total timesteps                     & $1.5\times10^7$           & $5\times10^6$             & $5\times10^6$              & $1\times10^7$              \\
Batch size                          & 8\,192                     & 16\,384                    & 2\,048                      & 2\,048                      \\
Mini-batch size                     & 1\,024                     & 2\,048                     & 256                         & 256                         \\
\# Environments                      & 16                         & 8                          & 16                          & 16                          \\
\# Steps / env                       & 512                        & 2\,048                     & 128                         & 128                         \\
Update epochs                       & 4                          & 10                         & 4                           & 4                           \\
Discount factor ($\gamma$)          & 0.995                      & 0.99                       & 0.99                        & 0.99                        \\
GAE $\lambda$                       & 0.95                       & 0.95                       & 0.95                        & 0.95                        \\
Learning rate (Adam)                & $2.5\times10^{-4}$         & $3\times10^{-4}$           & $2.5\times10^{-4}$          & $2.5\times10^{-4}$          \\
Value loss coef.\ ($c_v$)           & 0.5                        & 0.5                        & 0.5                         & 0.5                         \\
Clip coefficient                    & 0.1                        & 0.1                        & 0.1                         & 0.1                         \\
Max grad norm                       & 0.5                        & 0.5                        & 0.5                         & 0.5                         \\
\bottomrule
\end{tabular}
\end{table}

\begin{table}[h!]
\centering
\scriptsize
\setlength{\tabcolsep}{2pt} 
\caption{Model-Specific Hyperparameters (values listed as MiniGrid / MuJoCo / Classic / Atari)}
\label{tab:model_specific_hyperparams}
\begin{adjustbox}{width=1.\textwidth,center}
\begin{tabular}{@{}lccccccc@{}}
\toprule
\textbf{Param}          & \textbf{PPO-4 / PPO-1} & \textbf{LSTM} & \textbf{GRU} & \textbf{TRXL} & \textbf{GTrXL} & \textbf{Mamba} & \textbf{Mamba-2} \\
\midrule
Hidden dim              & 512/90/256/512   & 512/64/256/512   & 512/64/256/512   & 384/64/284/512   & 376/64/336/448   & 380/70/284/450   & 384/64/164/512   \\
Entropy coef.           & 1e-4/0/0.01/0.01 & 1e-4/0/0.01/0.01 & 1e-4/0/0.01/0.01 & 1e-2/0.01/0.01/0.01 & 0.01/0.01/0.01/0.01 & 1e-2/0/0.01/0.01 & 1e-2/0/0.01/0.01 \\
RNN hidden dim          & --                & 256/64/128/256   & 256/78/160/256   & --                & --                & --                & --                \\
TRXL layers             & --                & --                & --                & 3/3/1/1           & 2/2/2/1           & --                & --                \\
TRXL heads              & --                & --                & --                & 4/4/4/2           & 4/4/4/2           & --                & --                \\
TRXL memory len.        & --                & --                & --                & 119/64/32/64      & 119/64/32/64      & --                & --                \\
$d_\text{state}$        & --                & --                & --                & --                & --                & 64/64/64/64       & 128/64/128/64     \\
$d_\text{conv}$         & --                & --                & --                & --                & --                & 4/4/4/4           & 4/4/4/4           \\
Expand ratio            & --                & --                & --                & --                & --                & 2/1/2/1           & 2/2/2/1           \\
Learning rate           & --                & --                & --                & --                & --                & 1.5e-4/3e-4/1.5e-4/1.5e-4 & 1.5e-4/3e-4/1.5e-4/1.5e-4 \\
\bottomrule
\end{tabular}
\end{adjustbox}
\end{table}

\clearpage
\newpage

\subsection*{Licenses for Existing Assets}
We use the following open‐source assets, with licenses listed for each:
\begin{itemize}[noitemsep,leftmargin=*]
  \item \textbf{gym-minigrid (MiniGrid)}: Apache License 2.0.
  \item \textbf{MuJoCo physics engine}: Apache License 2.0; \textbf{mujoco-py} bindings: MIT License.
  \item \textbf{Arcade Learning Environment (Atari ALE)}: GNU General Public License v3.0.
  \item \textbf{CleanRL}: MIT License.
  \item \textbf{mamba-ssm}: MIT License.
\end{itemize}


\end{document}